\documentclass[conference]{IEEEtran}
\IEEEoverridecommandlockouts
\usepackage{cite}
\usepackage{amsmath,amssymb,amsfonts}
\usepackage{algorithm}
\usepackage{algpseudocode}
\usepackage{graphicx}
\usepackage{textcomp}
\usepackage{xcolor}

\usepackage{arydshln}  
\usepackage{multirow}  

\definecolor{lightblue}{RGB}{173, 216, 230}
\definecolor{gold}{RGB}{255, 215, 0}
\def\BibTeX{{\rm B\kern-.05em{\sc i\kern-.025em b}\kern-.08em
    T\kern-.1667em\lower.7ex\hbox{E}\kern-.125emX}}

\begin{document}

\title{Weakly-Supervised Domain Adaptation \\with Proportion-Constrained Pseudo-Labeling
\thanks{This study is supported by Grant-in-Aid for Challenging Research (Exploratory) JP23K18509, Cross-ministerial Strategic Innovation Promotion Program (SIP) on ``Integrated Health Care System'' Grant Number JPJ012425, and JST ASPIRE Program Japan Grant Number JPMJAP2403.} 
}


\author{
Takumi Okuo\IEEEauthorrefmark{1},
Shinnosuke Matsuo\IEEEauthorrefmark{1},
Shota Harada\IEEEauthorrefmark{1},
Kiyohito Tanaka\IEEEauthorrefmark{2}
Ryoma Bise\IEEEauthorrefmark{1} \\
\IEEEauthorblockA{\IEEEauthorrefmark{1}Kyushu University, Fukuoka, Japan, \IEEEauthorrefmark{2}Kyoto Second Red Cross Hospital, Kyoto, Japan} 
\texttt{\{takumi.okuo@human., shinnosuke.matsuo@human., harada@, bise@ \}ait.kyushu-u.ac.jp}
}

\maketitle

\begin{abstract}
Domain shift is a significant challenge in machine learning, particularly in medical applications where data distributions differ across institutions due to variations in data collection practices, equipment, and procedures. This can degrade performance when models trained on source domain data are applied to the target domain. Domain adaptation methods have been widely studied to address this issue, but most struggle when class proportions between the source and target domains differ. In this paper, we propose a weakly-supervised domain adaptation method that leverages class proportion information from the target domain, which is often accessible in medical datasets through prior knowledge or statistical reports. Our method assigns pseudo-labels to the unlabeled target data based on class proportion (called proportion-constrained pseudo-labeling), improving performance without the need for additional annotations. Experiments on two endoscopic datasets demonstrate that our method outperforms semi-supervised domain adaptation techniques, even when 5\% of the target domain is labeled. Additionally, the experimental results with noisy proportion labels highlight the robustness of our method, further demonstrating its effectiveness in real-world application scenarios.
\end{abstract}


\section{Introduction}
Domain shift is an important issue in machine learning in which the distribution of data in the target domain (test data in practice) differs from that in the source domain (training data). This can lead to significant performance degradation in models trained on the source domain data when applied to the target domain. In medical datasets, domain shift is particularly prominent due to differences in data collection practices, imaging equipment, and procedures across different hospitals. These differences can lead to variations in input data distributions, which present challenges for models trained on source domain to generalize effectively to the target domain.

Domain adaptation has been widely studied to address domain shift by transferring knowledge learned from a source domain with labeled data to a target domain with limited or no labeled data. This is crucial in applications like medical imaging, where labeled data is often scarce or expensive to obtain. Previous studies in domain adaptation have focused mainly on unsupervised domain adaptation (UDA)\cite{ganin2016domain,long2016unsupervised,ADDA,kang2019contrastive}, where no labels are available in the target domain, and semi-supervised domain adaptation (SSDA)\cite{MME,CDAC,SLA,SSSD}, where a small number of labeled samples are available. These approaches aim to align the data distributions between the source and target domains, such that the source and target data distributions become similar, making models more robust to domain shifts.

\begin{figure}[t]
    \centering
    \includegraphics[width=\linewidth]{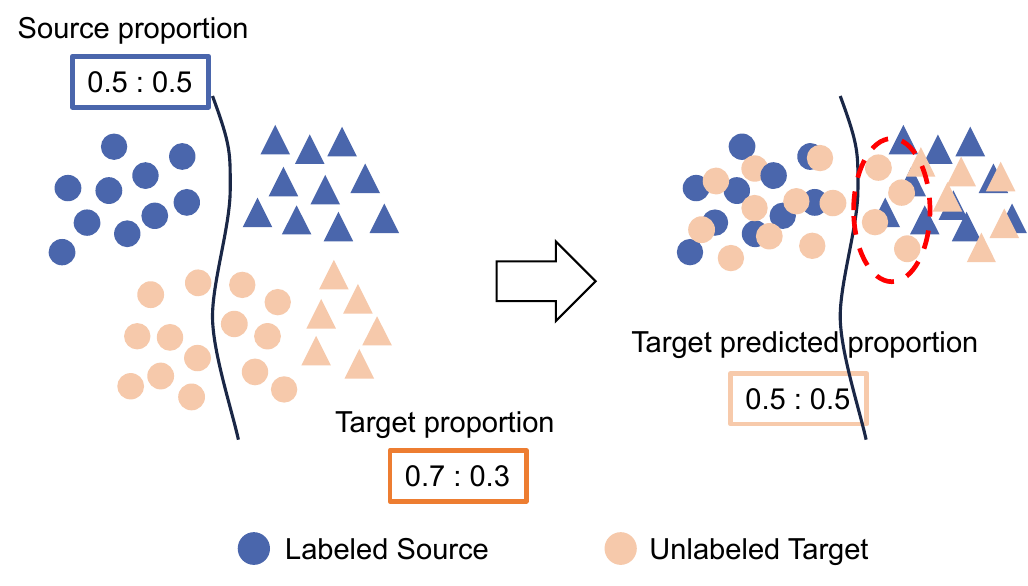}
    \caption{An example of class proportion mismatch between the source and target domains highlights that, in existing domain adaptation methods, even when the distributions of the source and target domains are aligned, decision boundaries may still become inconsistent.}
    \label{fig:introduction}
\end{figure}

Despite advances in domain adaptation, several important challenges remain. One major issue is the difficulty of improving performance when there is a mismatch in class proportions between the source and target domains. Many existing domain adaptation methods focus on aligning the distributions of the source and target domains or assigning pseudo-labels to target data. However, these approaches are highly dependent on the class proportions of the source domain, making them susceptible to bias. As a result, significant mismatches in class proportions lead to models being biased towards the source domain, resulting in poor performance on the target domain. For example, as shown in Figure~\ref{fig:introduction}, we consider a binary classification task in which, in addition to the fundamental distribution shift inherent in domain adaptation, the class proportions also differ between domains. In this setting, the class ratio in the source domain is (0.5, 0.5), whereas in the target domain it is (0.7, 0.3). Even when the overall data distributions are aligned, the class-wise distributions may still remain mismatched.


To address this issue, we consider the scenario in which the class proportions of the target domain are known. In many medical datasets, while individual labels may not be available, statistical information about class proportions is often accessible. Specifically, the distribution of health conditions, such as the proportions of healthy, low-severity, medium-severity, and high-severity cases, is frequently known from prior knowledge, diagnostic reports, or epidemiological studies. We assume that this statistical information can be leveraged in domain adaptation to develop a method that performs well even when the class proportions differ between the source and target.

We propose a weakly-supervised domain adaptation method that adapts the classifier using a large amount of unlabeled data and the class proportions in the target domain. The proposed method assigns pseudo-labels to each unlabeled data by solving an assignment problem with class proportion constraints (called proportion-constrained pseudo-labeling). These pseudo-labels are then used to train a classifier in a fully supervised manner. Furthermore, the pseudo-labels are iteratively updated at each epoch based on the model's predictions.


We validated the proposed method using two endoscopic datasets and demonstrated its superior performance. Unlike existing semi-supervised domain adaptation approaches, which require additional annotations for 5\% (approximately 400 images) of the target domain, our method leverages class proportions—a statistical property—without requiring extra annotations. Furthermore, it achieved higher accuracy than semi-supervised learning with the additional 5\% labeled data. To assess its robustness in practical scenarios, we also evaluated the method under inaccurate class proportions in the target domain and confirmed that it maintained high accuracy even when the statistical information was imprecise.

The main contributions of this paper are as follows:
\begin{enumerate}
    \item We propose a novel weakly-supervised domain adaptation problem setup that leverages the class proportions of the target domain, which is often accessible in medical datasets through prior knowledge or statistical reports.
    \item We propose proportion-constrained pseudo-labeling to address domain adaptation in cases where class proportions differ between domains.
    \item The proposed method achieves higher accuracy compared to current semi-supervised learning methods that require an additional 5\% labeled data.
    \item We demonstrated that our method still achieved high accuracy even when the class proportions in the target domain were inaccurate, i.e., in real-world application scenarios.
not precise.
\end{enumerate}

\section{Related work}

\subsection{Domain adaptation (DA)}
Unsupervised domain adaptation~(UDA)~\cite{ganin2016domain,long2016unsupervised,ADDA,kang2019contrastive} aims to build a classifier for the target domain by using labeled source samples and unlabeled target samples. A famous approach is training that minimizes the distance between the distributions of the source and target domains, i.e., the domain gap~\cite{ADDA,kang2019contrastive,long2016unsupervised,saito2018maximum,na2021fixbi}. To reduce this gap, several methods have been proposed, including direct minimization of the distribution distance, such as maximum mean discrepancy, adversarial learning~\cite{ADDA}, and contrastive learning~\cite{kang2019contrastive}. Another approach leverages pseudo-labeling~\cite{pan2020unsupervised,Gu_2020_CVPR}. This approach is difficult because it involves assigning pseudo-labels to target samples while addressing the domain gap. However, unlike the distance minimization-based approach described earlier, this approach allows for class-wise distribution alignment.

\par
Semi-supervised domain adaptation~(SSDA) is a variant of UDA and could use a small amount of labeled target samples in training a model~\cite{MME,CDAC,SLA,SSSD}. Since SSDA achieves significant accuracy improvement compared to UDA with a few annotation costs, it is attracting attention as a practical strategy in fields with high annotation costs, such as the medical image analysis task. The basic approach of SSDA is similar to UDA; however, it has significant advantages, including the accurate estimation of the class-wise distribution in the target domain with labeled target samples. However, UDA methods and SSDA methods face inherent challenges in addressing adaptation difficulties caused by class imbalance disparities across domains. In contrast, our proposed approach leverages proportion information from the target domain through weakly-supervised domain adaptation to achieve improved accuracy.

\par
Weakly-supervised domain adaptation has been explored in several studies, where labeled data is available in the source domain, but only weak supervision is available in the target domain. Many of these studies focus on segmentation or detection tasks, where class labels for each image are used as weak labels~\cite{das2023weakly,yamazaki2021weakly,doan2024weakly,10.1007/978-3-031-50069-5_18}. In such tasks, domain adaptation is performed by using weak supervision to estimate segmentation masks or bounding boxes from class labels in the target domain. Another form of weak supervision involves noisy labels~\cite{tan2019weakly,9428071,yu2021divergence,shu2019transferable}. In these tasks, the source domain labels contain noisy labels. However, these approaches are specifically designed for certain types of weak supervision and thus cannot be directly applied to our setup, where class proportions are available.

\subsection{Learning from label proportions (LLP)}
LLP~\cite{ardehaly2017co,tsai2020learning,pmlr-v157-yang21b,matsuo2023learning,ijcai2021p377,zhang2022learning,asanomi2023mixbag,kubo2024theoretical} is a weakly-supervised learning problem, where the goal is to obtain a classifier for individual samples given sets of samples, called bags, along with their corresponding class label proportions. Note that class labels for individual samples are not provided.
Recent research on LLP~\cite{ardehaly2017co,tsai2020learning,asanomi2023mixbag,kubo2024theoretical,matsuo2024lplp} commonly utilized proportion loss~\cite{ardehaly2017co}, which minimizes the distance between the given label proportion and the predicted proportion, computed as the average of the predicted probabilities of the samples within each bag.
Many methods extend proportion loss by incorporating regularization terms~\cite{tsai2020learning}, introducing pre-training~\cite{pmlr-v157-yang21b}, or applying bag-level data augmentation~\cite{asanomi2023mixbag}. Furthermore, several studies~\cite{matsuo2023learning,ijcai2021p377} have proposed pseudo-labeling-based LLP methods.
In our method, we treat the entire target dataset as a single bag and perform weakly-supervised domain adaptation by leveraging the class proportions of the target dataset.

\begin{figure*}[t]
    \centering
    \includegraphics[width=\linewidth]{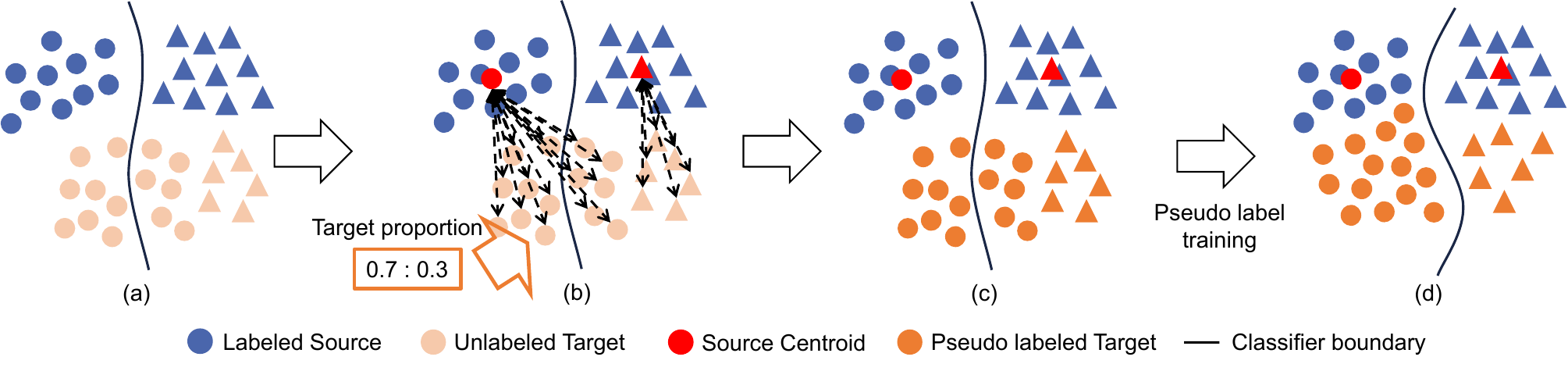}
    \caption{Overview of our method approach. (a) Training using only the source domain. (b) Calculating scores based on the distance between the source domain centroids and individual instances. (c) Assigning pseudo-labels by minimizing the scores under the constraint of satisfying class proportions. (d) Training using the pseudo-labels.}
    \label{fig:overview}
\end{figure*}

\begin{table}[t]
  \centering
    \caption{\textnormal{Preliminary experiments. The macro F1 scores of two methods: Pre-training only (before applying domain adaptation) and ADDA.}}
    \label{tab:preliminary}
    \resizebox{\linewidth}{!}{%
    \begin{tabular}{c|cc} \hline
        & Same proportion & Different proportion\\ \hline
        Pre-training only & 0.63 & 0.60\\
        ADDA~\cite{ADDA} & 0.79 & 0.63 \\
        \hline
        Improvement &  +0.16 & +0.03 \\
        \hline
    \end{tabular}}
\end{table}    

\section{Preliminary experiments}
In this section, we conduct preliminary experiments to demonstrate that domain adaptation becomes more challenging when the label proportions differ between domains.
In the experiments, we used SVHN as the source domain and MNIST as the target domain. We applied a widely used unsupervised adversarial domain adaptation method in two scenarios: one where the label proportions are the same between the source and target domains (same), and another where they differ (different).
Specifically, in the case of the same proportion, the label proportions are balanced (0.1, ..., 0.1) in both domains. In the case of the different label proportions, the source domain has a balanced label proportion (0.1, ..., 0.1), while the target domain has an imbalanced label proportion: (0.018, 0.036, 0.055, 0.073, 0.091, 0.109, 0.127, 0.145, 0.164, 0.182).

In the experiment, we evaluated how a simple unsupervised domain adaptation method can improve the macro-F1 (mF1) score compared to the 'Pre-training only' model, which is pre-trained using only the source domain. For domain adaptation, we used ADDA \cite{ADDA}, a widely used adversarial unsupervised domain adaptation method.

Table~\ref{tab:preliminary} shows the macro F1 scores of two methods: Pre-training only (before applying domain adaptation) and ADDA, along with the improvement achieved by ADDA. In the case where the label proportions are the same in both the source and target domains, ADDA improved mF1 significantly (+0.16). However, the improvement is limited in the case of different label proportions (only +0.03). This shows the difficulty of domain adaptation in the case of different label proportions between the source and target domains. This demonstrates the difficulty of unsupervised domain adaptation when the label proportions differ between the source and target domains. Many unsupervised domain adaptation methods focus on aligning the entire distribution between domains. However, even when the overall distribution is aligned, the class distributions (i.e., the density of each class) may still differ. Aligning these class distributions is challenging without additional information.

\section{Proposed method}
\subsection{Problem setting and overview}
Two datasets are given: the labeled source domain dataset $\mathcal{D}_S = \{ (x_S^{(i)}, Y_S^{(i)}) \}_{i=1}^{N_S}$ and the unlabeled target domain dataset $\mathcal{D}_T = \{ (x_T^{(j)}) \}_{j=1}^{N_T}$. Here, $x$ denotes a sample, $Y$ denotes the corresponding class label. $N_S$ and $N_T$ represent the number of samples of the source and target, respectively. Both domains share the same label space $y \in \{1, \dots, K\}$. Additionally, there exists a label proportion for the target data $\mathcal{D}_T$, denoted as $p_c$, which is expressed as $p_c = \frac{| \{ j \mid j \in \{1,\dots, N_T\}, Y_{c,j} = 1 \} |}{N_T}, (c=1, \dots, C)$. 
Here, $C$ is the number of classes, $Y_j$ indicates the one-hot vector about the class of $j$-th sample, and $|\cdot|$ indicates the cardinality of an input set. If the $j$-th sample belongs to class $c$, $Y_{c,j}=1$; otherwise, $Y_{c,j}=0$.
Note that the instance class labels $Y_j$ in the target domain are unknown, and only the class proportions are known.
The objective task is to train an instant classifier for the target domain using $\mathcal{D}_S$, $\mathcal{D}_T$, and $p_c, (c=1,\dots,C)$.

The proposed method consists of a two-step pipeline. First, supervised learning is performed using only the source domain data to pre-train the feature extractor $f_S$ and classifier $g_S$. Next, using $f_S$, $g_S$, and the feature distribution of the source domain, we perform proportion-constrained pseudo-labeling, which assigns a class to each sample (pseudo label), ensuring that the label proportion of the pseudo labels matches the predefined proportion $p_c$.

In the pre-training phase, supervised learning is conducted using the source domain data to enhance the classification accuracy within the source domain.
Specifically, we train $f_S$ and $g_S$ using the labeled source data $\mathcal{D}_S$ by the cross-entropy loss. 
The pre-trained network will be used for the next step.

\begin{algorithm}[tp]
\caption{Weakly-supervised domain adaptation with proportion-constrained pseudo-labeling} \label{alg}
\begin{algorithmic}[1]
    \State {\bf Inputs:}
    \Statex \hspace{0.5cm} Feature extractor $f_S$ and classifier $g_S$ trained
    \Statex \hspace{0.5cm} by source dataset, source centroid ${\mu}_1, \dots, {\mu}_C$
    \State {\bf Outputs:}
    \Statex \hspace{0.5cm} Trained feature extractor $f_T$ and classifier $g_T$
    \State {\bf Initialize:} 
    \Statex \hspace{0.5cm} Target feature extractor $f_T \gets f_S$
    \Statex \hspace{0.5cm} Target classifier $g_T \gets g_S$
    \For{epochs $t = 1, \dots, N_{\rm epoch}$}
        \State 1. Obtain the features $f_T(x_T^{(j)}) \quad \forall j \in \{1, \dots, N_T\}$
        \State 2. Assign the pseudo labels $\hat{Y}$ by Eq.(\ref{eq:mip})
        \State 3. Train $f_T, g_T$ using the pseudo labels $\hat{Y}$
    \EndFor
\end{algorithmic}
\end{algorithm}

\subsection{Proportion-constrained pseudo-labeling}
In this section, we describe the assignment of pseudo-labels constrained by the class proportion of the target dataset.
Simply assigning each target sample to the nearest class centroid of the source domain may lead to misclassifications due to the domain shift between the source and target domains. However, it is often the case that the relative positioning of class distributions within the target domain is similar to that in the source domain. Even if there is a distribution shift between the source and target domains, the assignment can be adjusted to better reflect the relative distribution within the domain, leading to a more appropriate assignment according to the proportions, as shown in Fig.~\ref{fig:overview}.

Based on this assumption, we solve an optimization problem that assigns pseudo labels $\hat{Y} \in \{0, 1\}^{C \times N_T}, \forall j \in \{1, \ldots, N_T\}, \sum^C_{c=1} Y_{c,j}=1$ while minimizing the distance between the assigned pseudo labels and the corresponding class centroids of the source dataset while preserving the class proportions.

Specifically, we solve the following optimization problem:
\begin{align}
\label{eq:mip}
&\min_{\hat{Y} \in \{0, 1\}^{C \times N_T}} \sum_{j=1}^{N_T} \sum_{c=1}^C \hat{Y}_{c,j} d_{c,j} \\
&\hspace{2mm} \text{s.t.} \ \sum^C_{c=1} \hat{Y}_{c,j}=1, \forall j \in \{1, \ldots, N_T\}, \nonumber \\
&\hspace{8mm} \sum^{N_T}_{j=1} \hat{Y}_{c, j} = p_c, \forall c \in \{1, \ldots, C\}, \nonumber
\end{align}
where $d_{c,j}$ represents the squared distance between the feature vector of the sample $x_T^{(j)}$ and the centroid $\mu_c$ of class $c$ in the source domain, which is computed as the mean of the feature vectors of samples belonging to class $c$ in the source domain.
The first constraint ensures that each sample is assigned to exactly one class, and the second constraint enforces that the number of assigned pseudo labels follows the given proportion.

The above optimization problem is an integer programming problem, which is generally NP-hard. However, the constraints can be represented by a totally unimodular constraint matrix. As a result, the problem can be relaxed to a linear programming (LP) problem, and the relaxed version can be solved in polynomial time.

\subsection{Training using pseudo labels on target dataset}
Algorithm~\ref{alg} shows the overview of the retraining process.
The networks for the target domain $g_T$, $f_T$ are first initialized using the pre-trained networks $g_S$, $f_S$.
Then, in each epoch, pseudo labels are assigned using proportion-constrained pseudo-labeling, and the parameters of the networks are updated with a cross-entropy loss using both the source data and the pseudo labels for the target data. Here, to mitigate the negative effects of class imbalance, oversampling is applied per batch. This process is iteratively repeated for each epoch until convergence on the validation set.

By leveraging both the source data and pseudo labels from the target domain, the network learns to adapt to the domain difference and adjusts the feature distributions extracted by the feature extractor $f_T$ to make them more similar across domains. Specifically, in the retraining step, we train the network using cross-entropy loss with pseudo-labels.


\section{Experiments}

\begin{table*}[h]
  \centering
    \caption{\textnormal{Classification performance on the endoscopic dataset. The evaluation indices are Accuracy, macro-Recall~(mRecall), macro-Precision~(mPrecision), macro-F1~(mF1). There are three types of approaches: not using target labels~(unlabel), using a small amount of target labels (semi), and using class proportions (proportion). Experiments were conducted in the case of "semi" where 1\%, 3\%, and 5\% of the entire target data were labeled.}}
    \label{Result_table}
    \resizebox{\linewidth}{!}{%
    \begin{tabular}{c|c|c|cccc|cccc} \hline
         &&&\multicolumn{4}{c|}{LIMUC to Private} & \multicolumn{4}{c}{Private to LIMUC}\\
        Type of target label & Target labels & Method & Accuracy & mRecall & mPrecision & mF1 & Accuracy & mRecall & mPrecision & mF1 \\ \hline \hline
        \multirow{2}{*}{Unlabel} 
         && Pre-training only & 0.267 & 0.382 & 0.324 & 0.218 & 0.561 & 0.332 & 0.384 & 0.289\\
         && ADDA\cite{ADDA} & 0.407 & 0.489 & 0.365 & 0.318 & 0.495 & 0.509 & 0.452 & 0.440\\ \hdashline
        \multirow{18}{*}{Semi} 
         & \multirow{6}{*}{1\%} & MME\cite{MME} & 0.610 & 0.560 & 0.484 & 0.469 & 0.670 & 0.537 & 0.558 & 0.541\\
         &                       & CDAC\cite{CDAC} & 0.588 & 0.546 & 0.446 & 0.446 & 0.593 & 0.539 & 0.502 & 0.506\\
         &                       & SLA\cite{SLA} & 0.586 & 0.549 & 0.460 & 0.455 & 0.650 & 0.480 & 0.546 & 0.489\\
         &                       & MME + SLA\cite{MME,SLA} & 0.589 & 0.547 & 0.476 & 0.464 & 0.670 & 0.543 & 0.568 & 0.546\\
         &                       & CDAC + SLA\cite{CDAC,SLA} & 0.596 & 0.551 & 0.464 & 0.454 & 0.606 & 0.536 & 0.507 & 0.503\\
         &                       & S$^{3}$D\cite{SSSD} & 0.653 & 0.580 & 0.494 & 0.507 & 0.666 & 0.519 & 0.541 & 0.522\\ \cdashline{2-11}
         & \multirow{6}{*}{3\%} & MME\cite{MME} & 0.588 & 0.591 & 0.468 & 0.476 & 0.679 & 0.559 & {\bf 0.605}& 0.570\\
         &                       & CDAC\cite{CDAC} & 0.670 & 0.592 & 0.524 & 0.512 & 0.607 & 0.573 & 0.527 & 0.532\\
         &                       & SLA\cite{SLA} & 0.582 & 0.560 & 0.468 & 0.471 & 0.664 & 0.534 & 0.577 & 0.540\\
         &                       & MME + SLA\cite{MME,SLA} & 0.599 & 0.604 & 0.477 & 0.487 & 0.688 & 0.576 & 0.604 & 0.582\\
         &                       & CDAC + SLA\cite{CDAC,SLA} & 0.665 & 0.621 & 0.510 & 0.522 & 0.614 & 0.578 & 0.527 & 0.536\\
         &                       & S$^{3}$D\cite{SSSD} & 0.674 & 0.623 & 0.519 & 0.537 & 0.677 & 0.578 & 0.584 & 0.567\\ \cdashline{2-11}
         & \multirow{6}{*}{5\%} & MME\cite{MME} & 0.608& 0.626 & 0.491 & 0.496 & 0.697 & 0.595 & 0.602 & 0.595\\
         &                       & CDAC\cite{CDAC} & 0.651 & 0.632 & 0.512 & 0.522 & 0.633 & 0.600 & 0.557 & 0.562\\
         &                       & SLA\cite{SLA} & 0.612 & 0.609 & 0.488 & 0.499 & 0.675 & 0.543 & 0.577 & 0.542\\
         &                       & MME + SLA\cite{MME,SLA} & 0.656 & {\bf 0.639}& 0.519 & 0.527 & {\bf 0.703}& 0.603 & 0.601 & 0.595\\
         &                       & CDAC + SLA\cite{CDAC,SLA} & 0.656 & 0.636 & 0.518 & 0.531 & 0.638 & 0.605 & 0.583 & 0.557\\
         &                       & S$^{3}$D\cite{SSSD} & 0.668 & 0.629 & 0.518 & 0.534 & 0.692 & 0.593 & 0.600 & 0.586\\ \hdashline
        \multirow{3}{*}{Proportion} 
         && proportion loss\cite{ardehaly2017co} & 0.630 & 0.502 & 0.412 & 0.398 & 0.603 & 0.462 & 0.443 & 0.404\\
         && ADDA + proportion loss\cite{ADDA,ardehaly2017co} & 0.413 & 0.475 & 0.356 & 0.309 & 0.559 & 0.534 & 0.493 & 0.485\\
         && Ours & {\bf 0.714}& 0.618 & {\bf 0.537}& {\bf 0.559}& 0.666 & {\bf 0.629}& 0.581 & {\bf 0.599}\\
        
        \hline
    \end{tabular}}
\end{table*}

\subsection{Dataset}
To demonstrate the effectiveness of our domain adaptation method for medical image recognition, we used endoscopic datasets consisting of two distinct domains; the Labeled Images for Ulcerative Colitis (LIMUC) dataset\cite{LIMUC} and the private dataset (Private).

LIMUC is a dataset of endoscopic images for Ulcerative Colitis (UC), which contains four classes based on the Mayo Endoscopy Score (MES) that reflect disease severity: Mayo 0 to 3, where a score of 0 indicates a healthy condition, and higher scores indicate increasing severity.
The sample distribution for each class in the LIMUC dataset was as follows: Mayo 0: 6105 samples (54.14\%), Mayo 1: 3052 samples (27.07\%), Mayo 2: 1254 samples (11.12\%), and Mayo 3: 865 samples (7.67\%). 
The Private dataset also consists of the same classes with LIMUC, and the sample distribution for the Private dataset was as follows: Mayo 0: 6678 samples (65.06\%), Mayo 1: 1995 samples (19.44\%), Mayo 2: 1395 samples (13.59\%), and Mayo 3: 197 samples (1.92\%).
Here, the label proportions differ between the domains, with the proportion of Mayo~3 being particularly different.


In the evaluation, we performed 5-fold cross-validation, where the dataset was divided into five parts, with one part used as the test set and the remaining parts used for training. For training on the target domain, we followed the current domain adaptation methods~\cite{MME} regarding validation data. Specifically, 10 labeled samples from each class were allocated as validation data, distinct from the test data. The remaining samples were used for training. The validation data were employed for early stopping during both pre-training and domain adaptation.

\subsection{Implementation details}
For the feature extractors and classifiers, the same structures were employed in both the pre-training phase on the source domain and the domain adaptation phase on the target domain. The feature extractor was constructed using a ResNet18\cite{Resnet} pre-trained on ImageNet, with its final fully connected layer removed. A linear classification layer was used for the classifiers $g_S$ and $g_T$, with the weights randomly initialized.


We used the Adam optimizer~\cite{Adam} with a learning rate of $1.0\times10^{-5}$ for pretraining and $1.0\times10^{-6}$ for retraining. The batch size was set to 64, and training was conducted for 100 epochs with early stopping (patience: 20). To address class imbalance during pretraining, oversampling was applied to equalize class distributions. Data augmentation included random rotations and horizontal/vertical flips. Since image sizes varied across domains, all images were resized to 228$\times$228 prior to input.


\begin{figure*}[t]
    \centering
    \includegraphics[width=\linewidth]{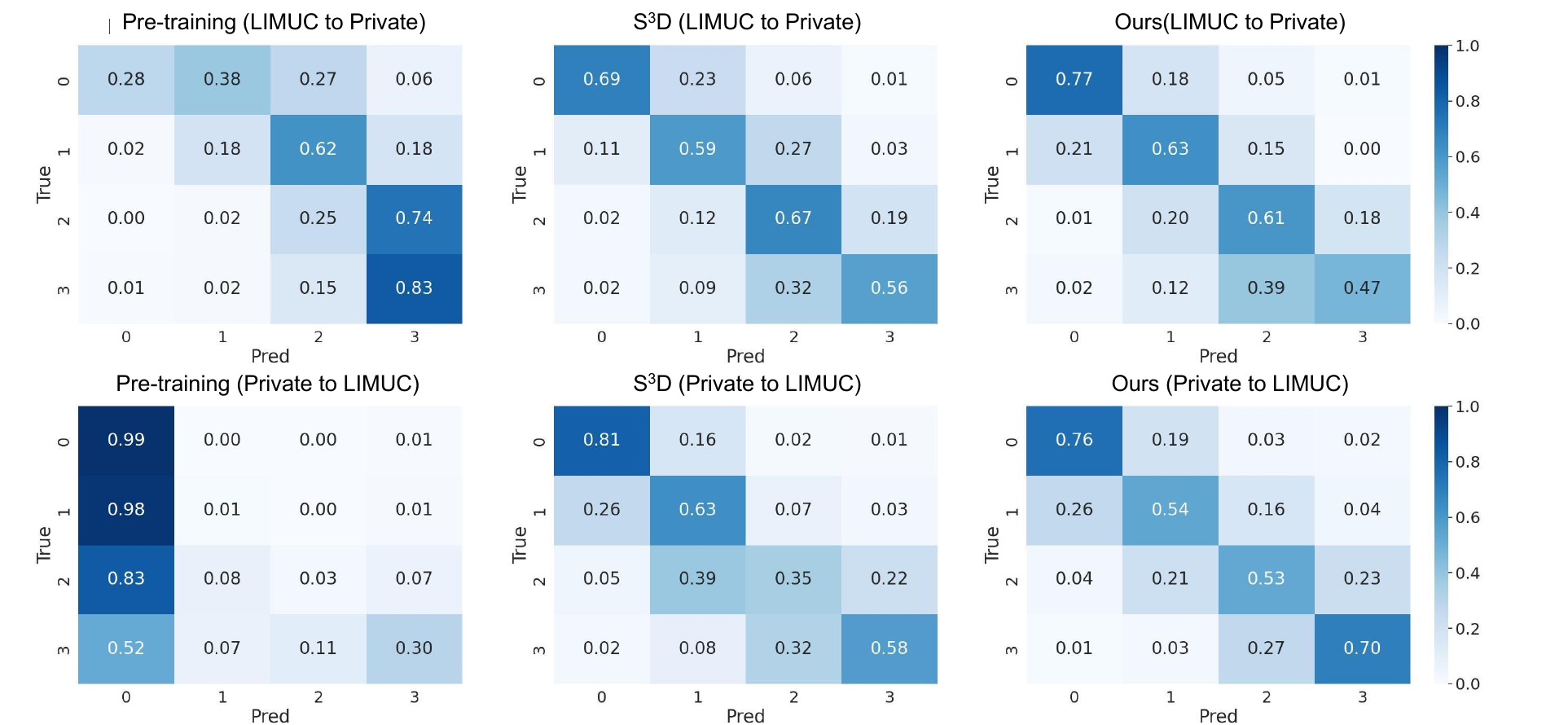}
    \caption{Confusion matrixs for pre-training only, S$^3$D, and Ours. The upper row represents domain adaptation from LIMUC to Private, while the lower row represents domain adaptation from Private to LIMUC. Row-wise normalization was applied, considering class imbalance.}
    \label{fig:matrix}
\end{figure*}
\subsection{Comparative methods}
We evaluated the performance of our domain adaptation method, which performs proportion-constrained pseudo-labeling. Since leveraging label proportions in the target domain for domain adaptation is a novel approach, there are no methods that directly address this setup. Instead, we compared our approach with methods that use proportion loss, UDA, and several semi-supervised domain adaptation (SSDA) methods:
1) Pre-train only, which is the model with pre-training using only the source domain;
2) ADDA\cite{ADDA}, which is an adversarial unsupervised domain adaptation;
3) MME\cite{MME}, which is a semi-supervised method and it minimizes the distributional difference between the source and target domains' feature distributions;
4) CDAC\cite{CDAC}, which is a semi-supervised method and it introduced an adversarial adaptive clustering loss;
5) MME+SLA\cite{MME,SLA}, which is a semi-supervised method and it introduces SLA into MME;
6) CDAC+SLA\cite{CDAC,SLA}, which is a semi-supervised method and it introduces SLA into CDAC;
7) S$^{3}$D\cite{SSSD}, which is a semi-supervised method and it uses self-distillation with sample pair;
8) Proportion loss\cite{ardehaly2017co}, which fine-tunes the pre-trained network using the proportion loss, where we trained the model by using the overall class proportions of the target domain as the ground truth label for all batches.
\par
For SSDA methods, we used 1\%, 3\%, and 5\% of the target domain's training data per class as labeled data.
While there may be slight variations across folds, the number of images used for each case is approximately 80, 240, and 400, respectively.
Comparing with SSDA is crucial to demonstrate the effectiveness of our method in reducing the number of additional annotations required for SSDA.

To evaluate the classification performance, we measured accuracy, the macro recall (mRecall), the macro precision (mPrecision), and the macro F1-score (mF1), which represent the averages of recall, precision, and F1 scores across all classes. Given the significant class imbalance within the dataset and the necessity of accurately classifying minority classes, mF1 is particularly critical as an evaluation metric.

\begin{figure*}[h]
    \centering
    \includegraphics[width=.95\linewidth]{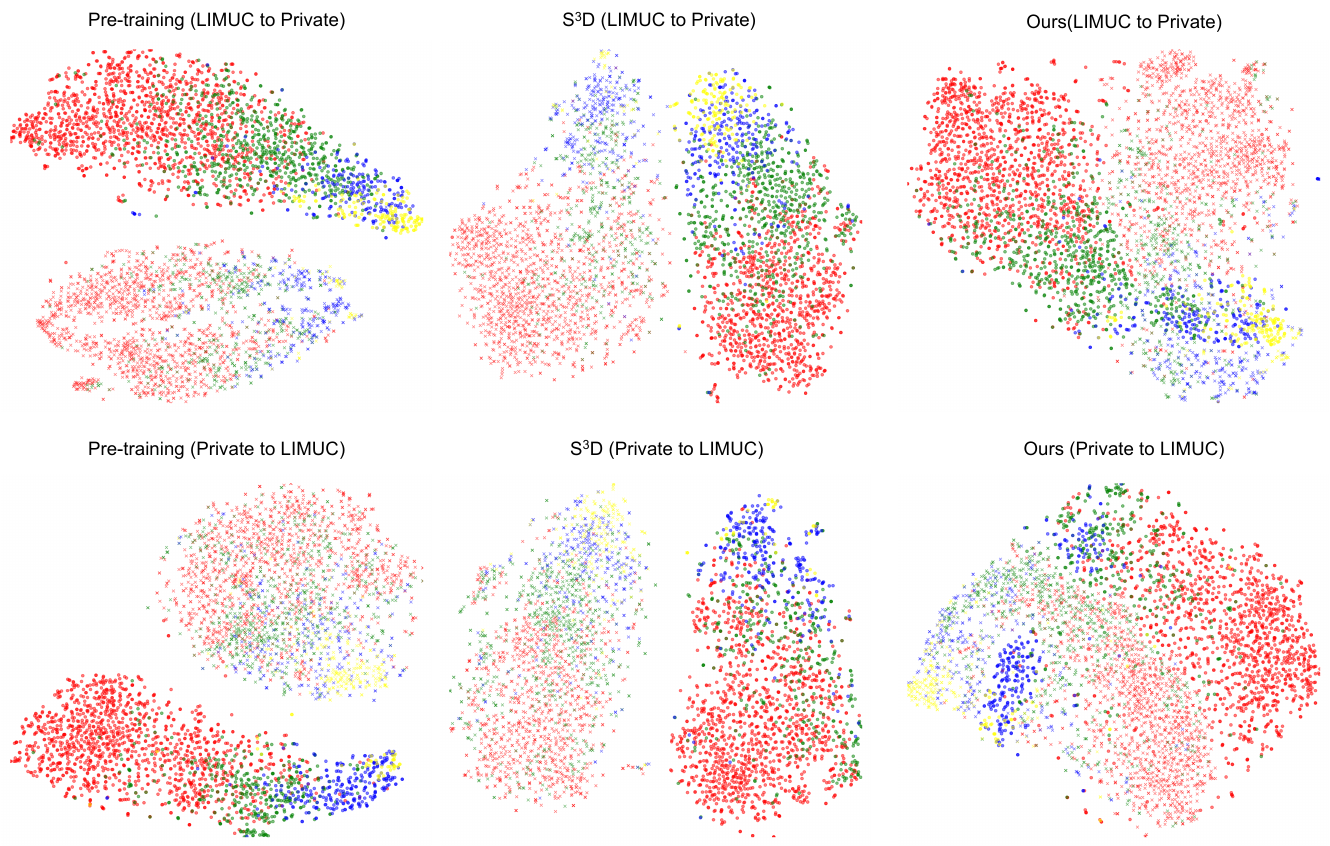}
    \caption{Feature visualization with t-SNE. Source domain is denoted by "\(\circ\)", target domain is represented by "\(\times\)". The colors indicate different categories: \textcolor{red}{red} for Mayo 0, \textcolor{green}{green} for Mayo 1, \textcolor{blue}{blue} for Mayo 2, and \textcolor{yellow}{yellow} for Mayo 3. The upper row represents domain adaptation from LIMUC to Private, while the lower row represents domain adaptation from Private to LIMUC.}
    \label{fig:TSNE}
\end{figure*}

Table~\ref{Result_table} shows the classification performances of the comparative methods under two scenarios: LIMUC as the source domain and Private as the target domain (LIMUC to Private), and vice versa (Private to LIMUC). In these scenarios, the performance of pre-training only is significantly worse due to the domain shift problem. Unsupervised domain adaptation with ADDA improves all performance metrics; however, the improvement is limited due to the difference in label proportions between the source and target domains. This domain adaptation task is a significant challenge for unsupervised domain adaptation approaches. Compared to these methods, our method significantly improved performance in both scenarios: LIMUC to Private and Private to LIMUC. Specifically, compared to ADDA, our method improved accuracy by +0.307 and mF1 by +0.241 in LIMUC to Private, and accuracy by +0.171 and mF1 by +0.159 in Private to LIMUC. These results demonstrate the effectiveness of utilizing proportion information for domain adaptation.

\begin{figure}[t]
    \centering
    \includegraphics[width=\columnwidth]{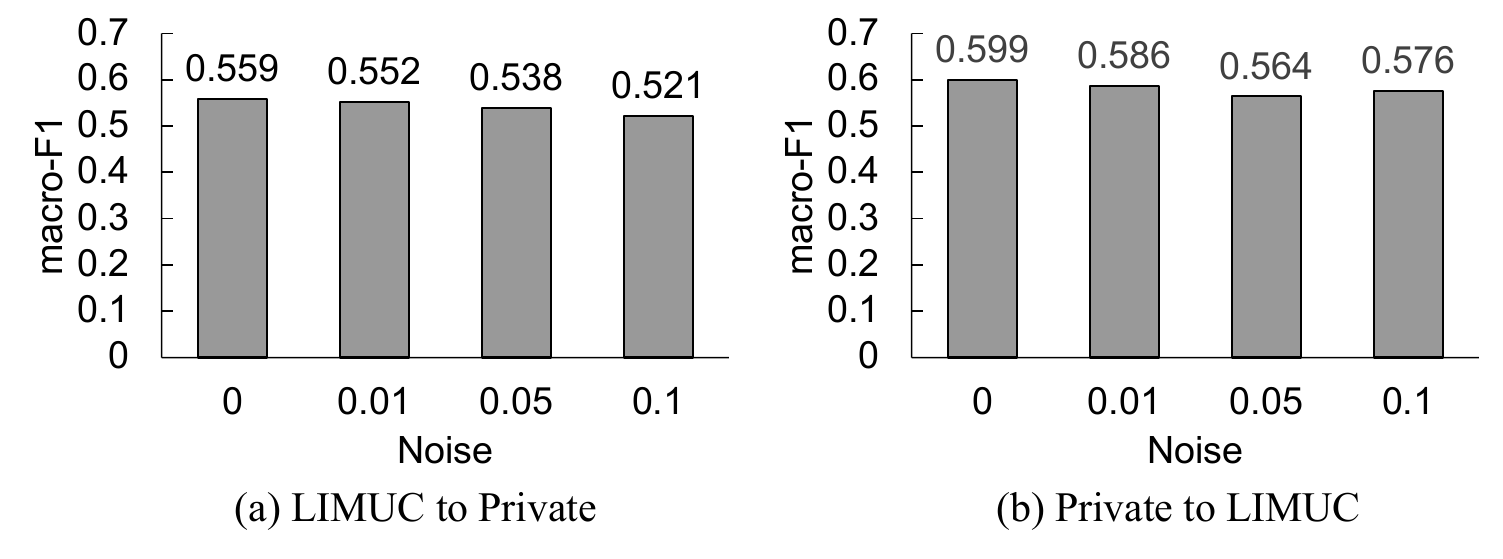}
    \caption{In cases where noise is present in the target domain's class proportions. Noise was introduced based on a Dirichlet distribution. Comparisons were conducted for scenarios where the total deviation from the true class proportions across all classes was set to 0.01, 0.05, and 0.1.}
    \label{fig:noise_result}
\end{figure}

In addition, our method outperformed the semi-supervised domain adaptation methods, which improved performance compared to ADDA by leveraging labeled data. While SSDA methods improved mF1 as the labeled data increased, our method outperformed these SSDA methods in terms of mF1 even when they used 5\% labeled data (about 400 images); our method was better than S$^{3}$D, which is the best in SSL in both scenarios. The results suggest that by utilizing label proportion information, which can be obtained as statistical information without additional annotations, our method achieves performance comparable to or better than adding annotations for 5\% (about 400 images).

ADDA + proportion loss, which simply incorporates the proportion loss commonly used in label proportion learning problems into ADDA, improved the mF1 score compared to pre-training only. However, the improvement is limited by such a simple extension using label proportion information. In contrast, our method significantly outperformed these approaches, demonstrating the effectiveness of our proportion-constrained pseudo-labeling for weakly-supervised domain adaptation using proportion information.

Fig. \ref{fig:matrix} presents the confusion matrices for three different methods: Pre-training only, S$^3$D (which uses 5\% labeled data and is the second best in terms of mF1), and our method, displayed from left to right. 
The rows of the confusion matrix represent the ground truth labels, and the columns represent the predicted labels. The matrices are row-normalized to account for class imbalances, with values expressed as proportions.

Pre-training tends to overestimate the higher severity classes in LIMUC to Private due to the domain shift and classifies almost all samples as Mayo 0 in Private to LIMUC. S$^3$D works well for LIMUC to Private, where it consistently improves recall for each class. However, in Private to LIMUC, the recall for Mayo 2 is not as good in S$^3$D. Our proposed method achieves improvements comparable to or better than S$^3$D, which requires 5\% labeled data, and performs relatively well in both scenarios. 

Fig. \ref{fig:TSNE} illustrates the feature distributions of these three methods in both scenarios, visualized using t-SNE\cite{t-SNE}. In Pre-training, the feature distributions of the source and target domains are completely separated in both scenarios. S$^3$D aligns the distributions based on the labeled data, though they remain fully separated in both scenarios. Although our method uses only the weak supervisory signal of class proportions, it effectively brings the source and target distributions closer together.

\subsection{Noise in target domain proportion}
In practical medical scenarios, situations may arise where the statistical information is imprecise, meaning that the given label proportion $\mathbf{p}$ may contain noise compared to the true distribution. To simulate experiments with noisy proportions, we randomly generated noise using a Dirichlet distribution, ensuring that the sum of the noisy proportion $\left| \mathbf{p} + \mathbf{\epsilon} \right|_1 = 1$. This approach has been used in previous studies to randomly determine the proportion of bags in label proportion-based learning tasks~\cite{zhang2022learning}. We evaluated three scenarios where the sum of the class-wise errors between the generated proportions and the true proportions, $\left| \mathbf{\epsilon} \right|_1 $, was set to 0.01, 0.05, and 0.1.

Fig. \ref{fig:noise_result} illustrates the results of the proposed method under conditions of noisy proportions. Generally, the accuracy decreased as the noise level increased. In terms of mF1, the difference between the highest and lowest performance was approximately 0.03 to 0.04 for both domain adaptations, indicating that minor noise does not significantly degrade performance.
In addition, although the performance slightly decreased due to the proportion noise, the mF1 scores remain comparable to those of the semi-supervised methods that used 3\% of labeled data. This demonstrates the effectiveness of our method in practical medical scenarios: when statistical information about class proportions is available, our method performs at a level comparable to SSL methods, without the need for additional annotations.
\section{Conclusion}

This paper proposed a novel weakly-supervised domain adaptation method with proportion-constrained pseudo-labeling, which assigns target samples to classes while satisfying the class proportions in the target domain. The proportion-constrained pseudo-labeling could obtain better pseudo-labels, even in scenarios where classification results tend to be biased, such as imbalanced classification tasks. Since class proportions are often accessible in many medical datasets, the proposed method can be adapted without additional annotation costs. Nevertheless, in experiments with the endoscopic image datasets, the proposed method outperformed several semi-supervised domain adaptation methods, which require additional annotation costs.

\section{Future work}
As demonstrated in this paper, it is likely that the label proportions provided for the target domain may not be accurate in real-world applications. In our experiments, we used the noisy proportion labels directly to select pseudo-labels, and observed that the mF1 score gradually decreases with the strength of the noise, which may lead to overfitting to the noisy label proportions. In future work, we aim to explore methods that take the noise distribution into account when selecting pseudo-labels. By incorporating such approaches, we hope to mitigate overfitting to noisy proportions and improve the robustness of the learning process.

In addition, we believe that utilizing the class proportion information of the target dataset can also be applied to semi-supervised domain adaptation. In future work, we aim to explore methods that can be integrated into various semi-supervised domain adaptation frameworks.
 \vspace{6mm}
 

\bibliographystyle{splncs04}
\bibliography{main}
\end{document}